\documentclass[a4paper,10pt,twocolumn]{ICCAS2024}
 
\usepackage{cite}
\usepackage{amsmath,amssymb,amsfonts}
\usepackage{graphicx}
\usepackage{float}
\usepackage{dblfloatfix}

\usepackage{balance}
\usepackage{mathtools}
\usepackage[colorlinks=true, urlcolor=blue, linkcolor=red]{hyperref}

\usepackage{diagbox}
\usepackage{booktabs} 
\usepackage{multirow} 

\begin{document}

\title{Deep Reinforcement Learning-based Quadcopter Controller: \\ A Practical Approach and Experiments}

\author{Truong-Dong Do${}^{1}$, Nguyen Xuan-Mung${}^{2}$ and Sung-Kyung Hong${}^{3*}$  }

\affils{
${}^{1}$Department of Aerospace System Engineering, Sejong University,\\
Seoul, 05006, South Korea ({${}^{1}$dongdo}@sju.ac.kr) \\
${}^{1,3}$Department of Convergence Engineering for Intelligent Drone, Sejong University,\\
Seoul, 05006, South Korea ({${}^{3}$skhong}@sejong.ac.kr) \\
${}^{2,3}$Faculty of Mechanical and Aerospace Engineering, Sejong University,\\
Seoul, 05006, South Korea ({${}^{2}$xuanmung}@sejong.ac.kr) 
{\small${}^{*}$ Corresponding author}}


\abstract{
   Quadcopters have been studied for decades thanks to their maneuverability and capability of operating 
   in a variety of circumstances. 
   However, quadcopters suffer from dynamical nonlinearity, actuator saturation, 
   as well as sensor noise that make it challenging and time consuming to obtain accurate dynamic models and achieve satisfactory control performance. 
   Fortunately, deep reinforcement learning came and has shown significant potential in system modelling and control of autonomous 
   multirotor aerial vehicles, with recent advancements in deployment, performance enhancement, 
   and generalization.
   In this paper, an end-to-end deep reinforcement learning-based controller for quadcopters 
   is proposed that is secure for real-world implementation, data-efficient, and free
   of human gain adjustments.
   First, a novel actor-critic-based architecture is designed to map the robot states 
   directly to the motor outputs.
   Then, a quadcopter dynamics-based simulator was devised to facilitate the training 
   of the controller policy.
   Finally, the trained policy is deployed on a real Crazyflie nano quadrotor platform, without any additional fine-tuning process.
   Experimental results show that the quadcopter exhibits satisfactory performance as it tracks a given complicated trajectory, which demonstrates the effectiveness and 
   feasibility of the proposed method and signifies its capability in filling the simulation-to-reality gap.
   }


\maketitle


\section{Introduction}
   Unmanned aerial vehicles (UAVs), in particular quadrotors, are increasingly 
   prevalent research subjects due to their flexibility and autonomy 
   \cite{do2023vision, tran2018vision, do2022multi}. 
   With hovering capabilities and vertical take-off and landing (VTOL), 
   a plethora of applications including search and rescue, infrastructure assessment, 
   and urban air mobility \cite{wang2023cooperative, lyu2023unmanned}.
   Precision and agility in flying movements are essential in these applications.
   The dynamics of quadrotors are highly nonlinear and often hard to 
   model specific system, which presents a considerable challenge in terms of 
   stabilization control. 
   Conventional cascaded hierarchy controls require domain expertise and engineering 
   to be adjusted for changing hardware and utilization circumstances. Moreover,
   this method often takes extensive setup and experiment time, along with 
   parameter optimization \cite{xuan2023novel, do2023efficient}. 

   Recent developments in machine learning and deep learning have proven efficient 
   in addressing various complicated problems \cite{kuutti2020survey, do2018real}.
   Robot control is a decision-making problem that can be characterized as a Markov 
   Decision Process (MDP), in contrast to supervised learning, in which labels tend 
   to be not directly exist.
   In order to deal with MDPs, reinforcement learning has been utilized to train 
   policies for complicated end-to-end control problems in simulation 
   \cite{lillicrap2015continuous, schulman2015high, schulman2017proximal}.
   Additionally, they demonstrate encouraging results in learning tasks 
   involving continuous state/action space \cite{lillicrap2015continuous}, which have 
   a close relationship to quadrotor control \cite{hwangbo2017control}.
   Compared with typical optimization methods, this technique eliminates the necessity 
   for a specified controller structure, which might constrain agent performance and 
   increase human effort. 
   Despite results achieved in simulation are incredible, the reality gap between 
   simulation and real-world systems is a notable challenge to the deployment of 
   trained control policies.
   This is essentially due to model imperfections, inaccurate state observations, 
   uncertainty from observation and action, and other disturbances 
   \cite{molchanov2019sim, huang2023datt}.

   In this research, an end-to-end reinforcement learning based quadcopter 
   controller is proposed to deal with the aforementioned problems.
   The aim is to create a controller that is extremely data efficient, 
   free of human gain adjustments, and is safe for real-world implementation.
   First, a novel actor-critic-based architecture is designed to 
   directly translate quadcopter states to motor Revolutions Per Minute 
   (RPM) outputs without any supplementary pre-configured controllers.
   Then, a python simulated environment was created using OpenAI Gym 
   \cite{brockman2016openai}, 
   specifically designed for training transferable policies. The setting 
   is based on a lightweight quadcopter platform known as Crazyflie 2.1\footnote{Crazyflie 2.1: https://www.bitcraze.io/products/crazyflie-2-1/} 
   \cite{giernacki2017crazyflie}, 
   which has been equipped with thrust upgraded motors and propellers.
   Finally, practical flying tests were conducted to validate the 
   adaptation of the sim2real strategy.
   The preliminary results indicate that using a neural network, trained 
   with reinforcement learning methods in a simulated environment, is 
   both effective and feasible for comprehensive control of the drone.

   The remainder of this article are organized in the following manner.
   Section \ref{Methodology} \ provides an overview of the methodologies 
   including the quadcopter dynamics simulated environment, policy 
   networks, and the training strategy.
   The experiments, involving the system's configuration and the 
   obtained results, are shown in section \ref{Experiments} \ 
   Furthermore, in section \ref{Conclusion} \ , paper concludes by 
   outlining potential future research and development opportunities.

\vspace{-3mm}
\section{Methodology}
\label{Methodology}
\begin{figure}[t!]
      \includegraphics[width=8.5cm]{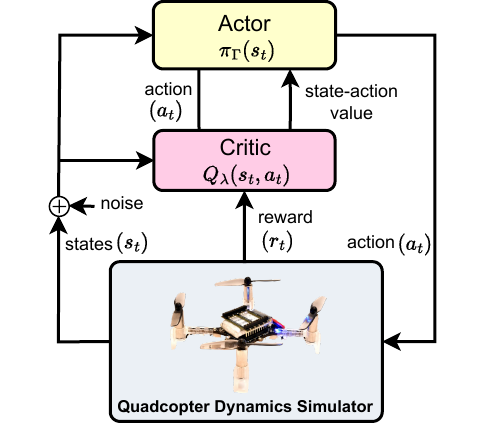}
      \vspace*{-2mm}
      \caption{Actor-critic-based end-to-end deep reinforcement learning quadcopter controller approach diagram.}
      \centering
      \label{fig:diagram}
   \end{figure}

   The diagram of end-to-end deep reinforcement learning quadcopter controller
   algorithm is demonstrated in Fig. \ref{fig:diagram}. 
   The quadcopter dynamics simulator is first established, followed by the 
   designing of an actor-critic-based policy network that directly maps the states 
   to the RPM commands of the motor.
   Quaternions are generally adopted to demonstrate orientation since they provide a global 
   and compact expression. However, there is a specific drawback in our case, which is the 
   existence of two figures that indicates the same rotation (i.e. $q = -q$). Therefore, we 
   must either increase the amount of training data by two-fold or accept the issue of 
   having a discontinuous function when we limit our domain to one hemisphere of $S^{3}$. 
   The rotation matrix is a representation that is highly redundant, but it is also simple 
   and free from such issues.
   Hence, in order to eliminate the ambiguity that rose from the quaternion's double 
   coverage of the space of rotations, we convert them to a nine-element rotation matrix $R$.

   Moreover, it is observed that the delay in the step-response is significant and is an effect of the low-pass behavior indicated by the motors.
   This behavior greatly affects the dynamics of the Crazyflie nano quadrotor used in 
   this study. According to the manufacturer's measurements\footnote{Crazyflie motor 
   step response: https://www.bitcraze.io/2015/02/measuring-propeller-rpm-part-3/},
   we have determined through empirical analysis that a time interval of $0.15$ seconds would be sufficient for 
   sim2real transference.
   These delays are considerably longer than the typical control interval of low-level 
   controllers in Crazyflie. Therefore, they result in actions that affect the state 
   only after $5$ to $25$ control steps \cite{eschmann2024learning}. In order to tackle the major amount of partial 
   observability, we include the history of control actions into the observation.
   This is a proprioceptive measurement that can be easily done in software without 
   the need of additional hardware.
   In addition, noise was introduced into the observation to imitate imperfections in 
   the sensor feedback of the reality platform \cite{eschmann2024learning}. The unexpected noise values have a 
   standard deviation of $0.001$ for position and orientation, and $0.002$ for linear 
   velocity and angular velocity, respectively.

   The observations are $18 + 4 \cdot N_{H_a}$ dimensional, where $N_{H_a}$
   is the number of action history \cite{eschmann2024learning}, described as follows:
   \begin{align}
      \label{eqn:state}
      s_t = [p_{t}, R_{t}, v_{t}, \omega_{t}, H_{a}]
   \end{align}
   where $p_{t}$ denotes the position;
   $R_{t}$ is the rotation matrix;
   $v_{t}$ is linear velocity;
   $\omega_{t}$ is the angular velocity;
   $H_{a}$ is the history of actions.

   The actions comprise the RPM setpoints for each motor inside of the 
   continuous space range of $[\mathrm{min}_{a}, \mathrm{max}_{a}]$ shown as below:
   \begin{align}
   \label{eqn:action}
      a_t = [r_{1}, r_{2}, r_{3}, r_{4}] = \pi_{\Gamma}(s_t)
   \end{align}

   \vspace{-3mm}
   \subsection{Quadcopter dynamics simulator}
   \label{Quadcopter_dynamics_simulator}
\begin{figure}[t!]
      \includegraphics[width=8.5cm]{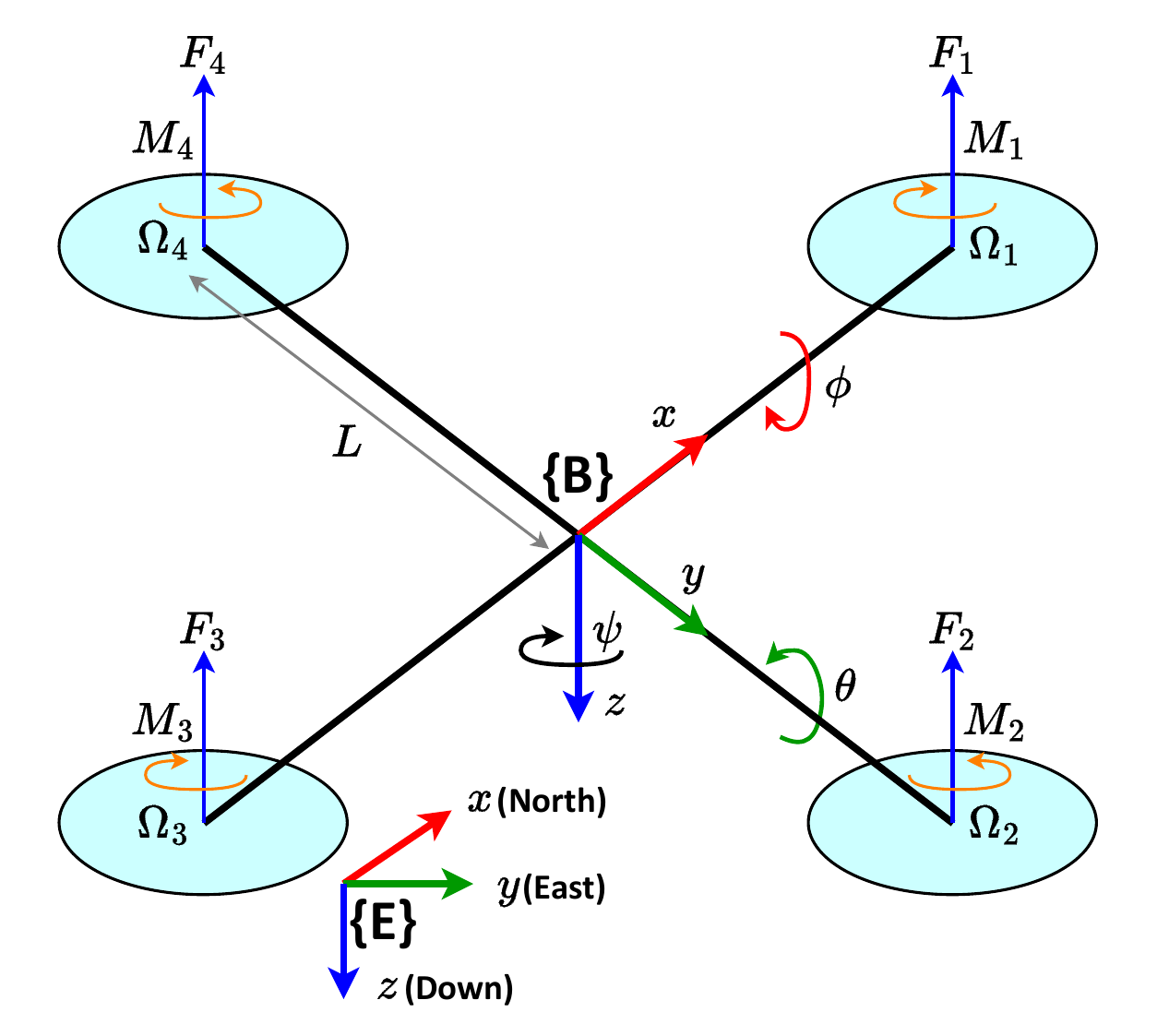}
      \vspace*{-2mm}
      \caption{The configuration of the quadcopter and coordinate systems with 
      a body frame \{B\} (x, y, z) and the Earth frame \{E\} (N, E, D).}
      \centering
      \label{fig:quad_config}
   \end{figure}

   We create a simulator that relies on the quadcopter dynamics model to train the 
   policy network.
   In previous studies, a number of methods have been proposed and demonstrated to 
   determine the dynamics model of the quadcopter using simulations and experiments 
   \cite{do2023efficient,xuan2023novel}.
   Let $ \Xi = [\phi,\theta,\psi]^{T} \in \mathbb{R}$$^{3}$ denotes the quadcopter's 
   attitude including roll, pitch, and yaw angles in the Earth frame \{E\} 
   as shown in Fig. \ref{fig:quad_config}.
   The vehicle's angular velocity in the body frame \{B\} denoted 
   as $\omega_{b} \in \mathbb{R}$$^{3}$.
   The position and velocity of the quadcopter along $\mathrm{x}$, $\mathrm{y}$, $\mathrm{z}$-axis of \{E\} 
   are indicated by $p = [x,y,z]^{T} \in \mathbb{R}$$^{3}$, 
   and $v = \dot{p}$, respectively.
   $J = \mathrm{diag}(I_{xx}, I_{yy}, I_{zz})$ $\in \mathbb{R}$$^{3 \times 3}$ denotes the moment of inertia in \{B\}. 
   The quadcopter's dynamics model can be described by the following equations:
   \begin{align}
      \label{eqn:quad_dynamic}
         \omega_{b} &= \mathrm{R}(\Xi) \dot{\Xi} \\
         m\dot{v} &= m g + \mathrm{R}(\Xi) u_{3} F_{\sum} \\
         J \dot{\omega}_{b} &= (J \omega_{b}) \times \omega_{b} + \tau 
   \end{align}
   where $m$ denotes the total mass of the quadcopter;
   $g \in \mathbb{R}$$^{3}$ is the gravitational acceleration vector; 
   $\mathrm{u}_{3}= [0,0,1]^{T}$;
   $F_{\sum}$ is the total thrust force; 
   $R(\Xi)$ is the rotation matrix from \{B\} to \{E\}, which is defined as:
   \begin{align}
      R(\Xi) = \begin{bmatrix}
         1 & 0            & \mathrm{-sin}(\theta) \\
         0 & \mathrm{cos}(\theta)  & \mathrm{sin}(\phi)\mathrm{cos}(\theta) \\
         0 & \mathrm{-sin}(\theta) & \mathrm{cos}(\phi)\mathrm{cos}(\theta)
      \end{bmatrix}
   \end{align}

   The control torque $ \tau = [\tau_{1}, \tau_{2}, \tau_{3}]^{T}$ is generated 
   by four propellers attached to four motors as follows:
   \begin{align} 
      \begin{cases}
         \tau_{1} = L C_{b}(\Omega_{2}^{2}-\Omega_{4}^{2})\cr
         \tau_{2} = L C_{b}(\Omega_{3}^{2}-\Omega_{1}^{2})\cr
         \tau_{3} = C_{d} C_{b}(-\Omega_{1}^{2}+\Omega_{2}^{2}-\Omega_{3}^{2}+\Omega_{4}^{2})
      \end{cases} 
   \end{align}
   where $L$ represents the arm length of the quadcopter;
   ${\Omega}_{i}$ denotes the $i$-th motor's rotary speed  $(i=1,2,3,4)$; 
   $C_{b}$ and $C_{d}$ consequently represent thrust and drag coefficient.

   \subsection{Networks architecture}
      We take advantage of the Twin Delayed Deep Deterministic policy gradient (TD3)
      \cite{fujimoto2018addressing}, an off-policy reinforcement learning approach 
      that provides enhanced sample complexity.

      There are two networks, specifically a value network and a policy network, 
      adopted for training. Both networks take the state as an input. 
      We involve a fully-connected neural network to represent a policy.
      The policy network is formed by two hidden layers, each containing $64$ neurons.
      The activation function employed in these layers is $\mathrm{tanh}$, whereas the output 
      layer uses linear activation. The value network has a same structure and is 
      trained for predicting the state-action value function.

   \subsection{Rewards}

      For the initial state distribution, we sample uniformly from the following sets:
      The position is sampled inside a $0.2 \mathrm{m}$ box centered around the origin location. 
      The orientation is represented by the rotation matrix $R$, which belongs to the
      entire $\mathrm{SO}(3)$ group and has a maximum angle of $90$ degrees. The linear
      velocity has a maximum magnitude of $1$ $\mathrm{m/s}$, while the angular 
      velocity has a maximum value of $1$ $\mathrm{rad/s}$. 
      In order to solve the issue of ``learning to terminate" \cite{eschmann2021reward}, 
      we use a negative squared reward function and includes an additional constant 
      to encourage survival as follows:
      \begin{align}
         \label{eqn:reward}
         r(s_{t},a_{t},s_{t+1}) = &\lambda_{s} - \eta_{p} \|p_{t}\|^{2}_{2} - \eta_{R} (1-R_{t}^{2}) \nonumber \\
                     &- \eta_{v} \|v_{t}\|^{2}_{2} - \delta_{a} \|a_{t} - \delta_{ab} \|^{2}_{2} 
      \end{align}
      where $\lambda_{s}$ is the survival bonus;
      $\eta_{p}$, $\eta_{R}$, $\eta_{v}$, and $\delta_{a}$ are the position,
      orientation, linear velocity, and action weights, respectively;
      $\delta_{ab}$ is the action baseline weight.

      The survival bonus is configured with a high value in order to 
      encourage the agent to maintain its life.
      The agent is penalized for deviating from the target state based 
      on its position, orientation, linear velocity, and action terms.
      The position and orientation components have the highest cost 
      coefficients due to their significant impact on the overall outcome.
      The other cost terms are chosen minimized to prevent the agent 
      from executing unreasonable actions, which might result in crashing 
      or straying from the desired state.

\vspace{-5mm}
\section{Experiments} 
\label{Experiments}
   \subsection{Systems Configurations}

   \begin{table}[b]
      \centering
      \caption{QUADCOPTER PARAMETERS USED FOR THE SIMULATOR.}
      \label{tb1}
      \resizebox{0.75\columnwidth}{!}{%
      \begin{tabular}{c c c}
         \toprule[1.5pt]
         \textbf{Parameter}           & \textbf{Value}           & \textbf{Unit}  \\ 
         \midrule[0.7pt]
         $m$                           & 0.033                     & kg \\ 
         $g$                           & $[0, 0, -9.81]^T$        & m/s$^2$ \\
         $L$                           & 0.028                    & m \\ 
         $C_{b}$                       & 0.0059                   & Ns$^{2}$ \\ 
         $C_{d}$                       & 9.18 $\times 10^{-7}$    & kg/rad \\
         $I_{xx}$                      & 16.57 $\times 10^{-6}$   & kgm$^{2}$ \\
         $I_{yy}$                      & 16.66 $\times 10^{-6}$   & kgm$^{2}$ \\
         $I_{zz}$                      & 29.26 $\times 10^{-6}$   & kgm$^{2}$ \\
         \bottomrule[1.5pt]
      \end{tabular}%
   }
   \end{table}

\begin{figure}[t!]
      \centering
      \includegraphics[width=7.5cm]{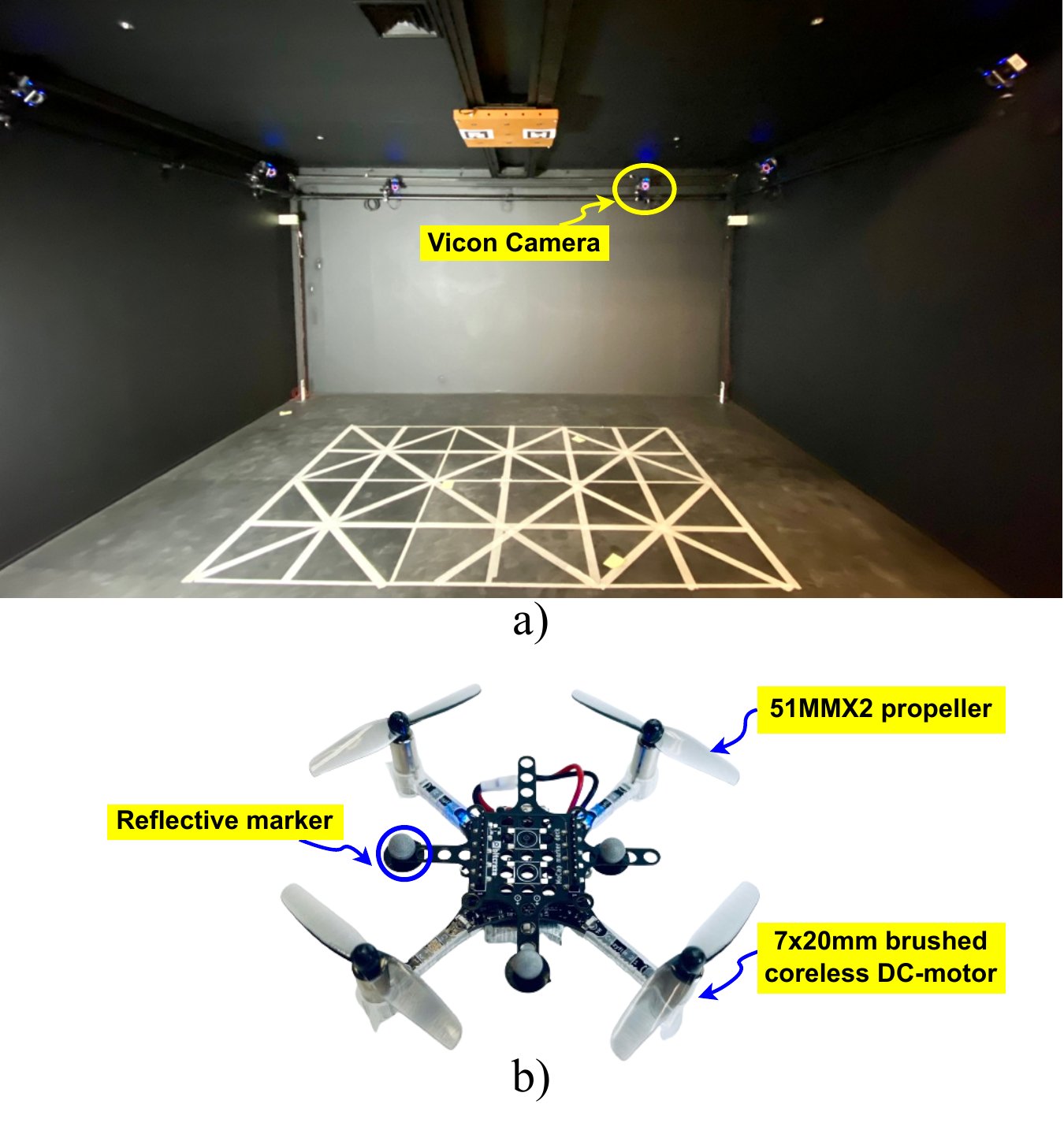}
      \vspace{-2mm}
      \caption{Experimental conditions. a) VICON positioning system, b) Crazyflie 2.1 with thrust upgrade kit}
      \centering
      \label{fig:exp_setup}
   \end{figure}

   We verify our approach with the Bitcraze Crazyflie 2.1 quadcopter, which is equipped with 
   the thrust improvement bundle package\footnote{https://www.bitcraze.io/2022/10/thrust-upgrade-kit-for-the-crazyflie-2-1/}. 
   This kit includes longer $7 \times 20 \mathrm{mm}$ brushed coreless DC-motors and a $51\mathrm{MM}$-$\mathrm{X}2$ propellers, 
   resulting in enhanced agility as seen in Fig. \ref{fig:exp_setup}b).
   The total weight of the assembled drone, which includes the batteries 
   and reflective markers, is only $33$ $\mathrm{grams}$.
   This platform has a thrust-to-weight ratio slightly lower than $2$ and
   utilizes an STM32F405 microcontroller operating at a clock speed of $168\mathrm{MHz}$.
   We use the pre-existing parameter estimates of the Crazyflie in \cite{cf_system2015} as 
   illustrated in the Table \ref{tb2}.

   The experiments were conducted in a $5 \times 4 \times 2.5 \mathrm{m}^{3}$ flying space at the 
   Guidance, Navigation and Control Lab (GNC Lab) at Sejong University with the VICON Motion 
   Capturing System consisting of $12$ cameras as illustrated in Fig. \ref{fig:exp_setup}a).
   It is assumed that we have access to actually precise estimates of the quadrotor's 
   position, orientation, linear velocity, and angular velocity. 
   The VICON system supplies the position and linear velocity data to a base station 
   computer at a frequency of $100 \mathrm{Hz}$, whereas the updates for orientation angle state 
   estimate occur at $1 \mathrm{kHz}$.
   The state estimate is obtained with an extended Kalman filter (EKF) \cite{kalman1960new} that combines data from
   the on-board IMU with motion capture information.

   The control policy is a C-function that is automatically generated from the trained 
   neural network model in PyTorch\footnote{https://pytorch.org/}. 
   Next, we included the functionality to output the RPMs of all four motors in the 
   Crazyflie firmware.
   The Crazyswarm API \cite{preiss2017crazyswarm} is used to handle communication with the drone. 
   The desired positions are sent to the quadrotor at a rate of $50 \mathrm{Hz}$ via 
   a $2.4 \mathrm{GHz}$ radio dongle.

   \subsection{Experimental results}
      The training is conducted with the TD3 implementation in the Stable Baselines3 
      library \cite{stable-baselines3}. 
      We train the policy for a total of $5$ million steps and takes slightly over $1$ hours 
      on a computer equipped Intel Core $\mathrm{i}9$-$13900\mathrm{K}$ processor and NVIDIA $4090$ GPU.
      The policy is trained with a batch size of $256$, a learning rate of $10^{-3}$.
      The length of the action history $N_{H_a}$  is set to $32$, the range action values is set as 
      $[\mathrm{min}_{a},\mathrm{max}_{a}]$=$[-21702,27102]$.
      The values of rewards function parameters are specified in Table \ref{tb2}.

      \begin{table}[b!]
         \centering
         \caption{PARAMETERS OF REWARDS FUNCTION.}
         \label{tb2}
         \resizebox{0.9\columnwidth}{!}{%
         \begin{tabular}{c c c}
            \toprule[1.5pt]
            \textbf{Parameter}    & \textbf{Value}         & \textbf{Description}    \\ 
            \midrule[0.7pt]
            $\lambda_{s}$        & 2            & Survival bonus           \\
            $\eta_{p}$           & 2.5          & Position weight          \\
            $\eta_{R}$           & 2.5          & Orientation weight       \\
            $\eta_{v}$           & 0.05         & Linear velocity weight   \\ 
            $\delta_{a}$         & 0.05         & Action weight            \\ 
            $\delta_{ab}$        & 0.35         & Action baseline          \\
            \bottomrule[1.5pt]
         \end{tabular}%
      }
      \end{table}
      \begin{figure}[ht!]
         \centering
         \vspace{-2mm}
         \includegraphics[width=8cm]{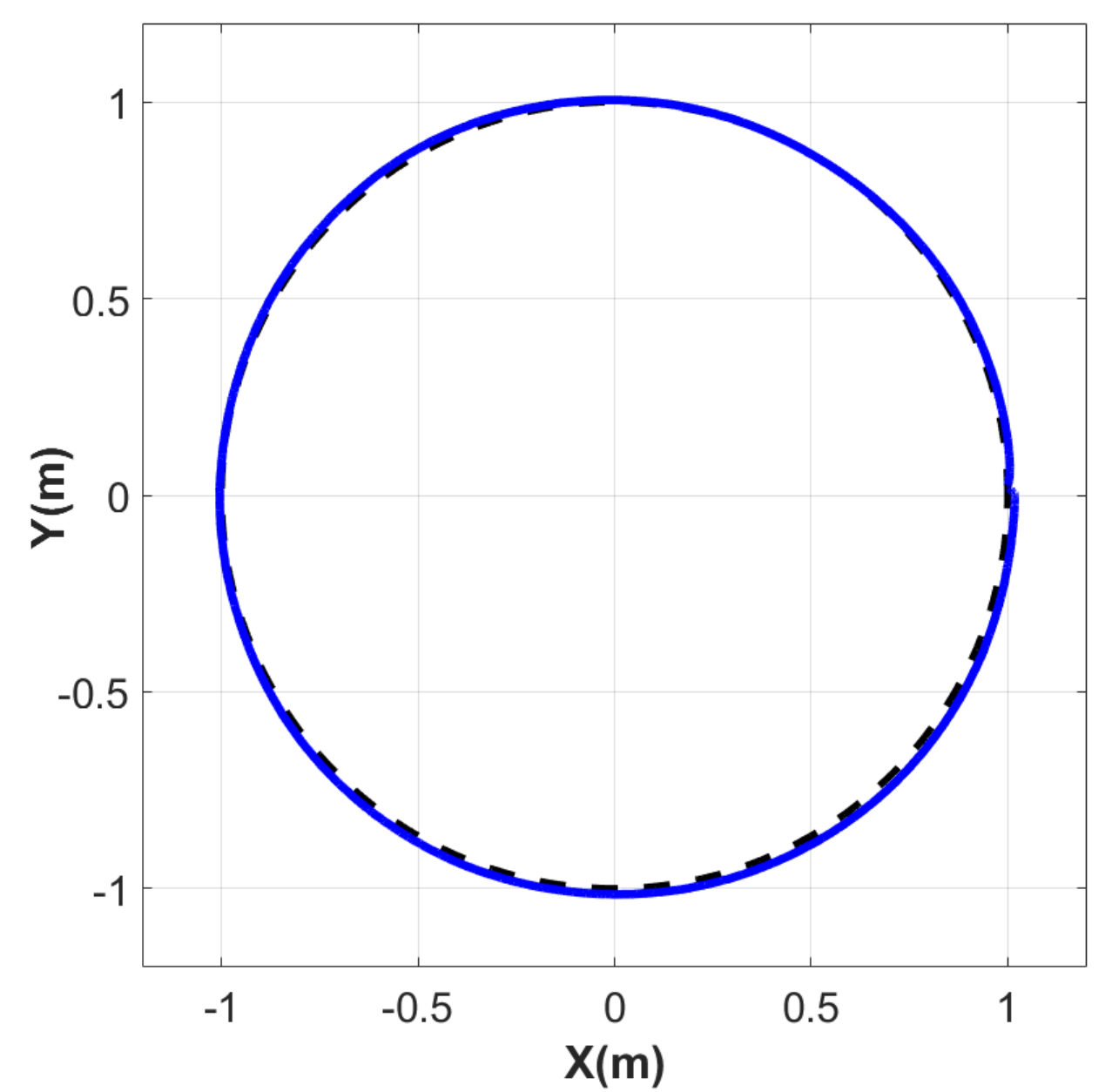}
         \caption{Real flight tracking performance of circle trajectory 
         with a radius of $1\mathrm{m}$ in $T$=$6$ seconds.}
         \label{fig:circle_trajectory}
      \end{figure}
      \begin{figure*}[t!]
         \centering
         \includegraphics[width=\textwidth]{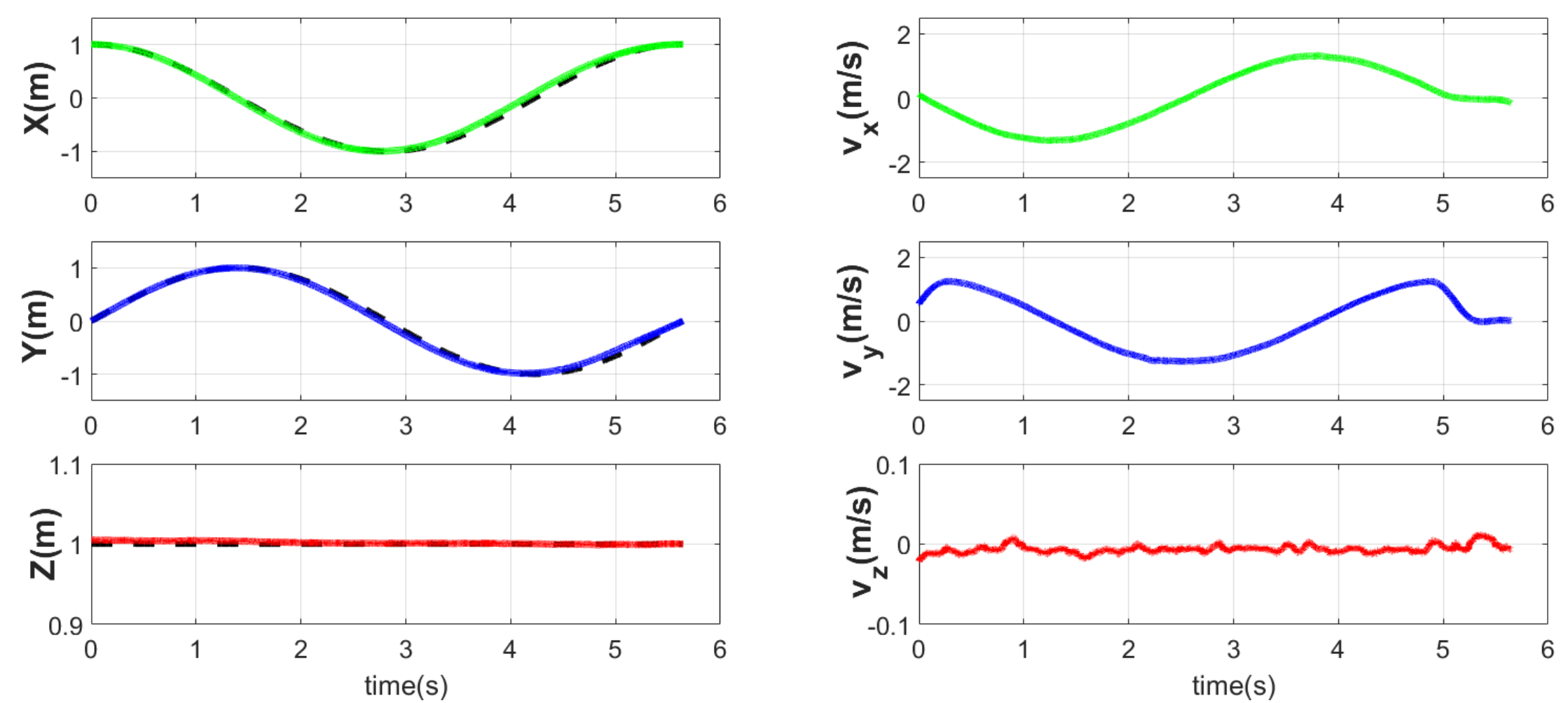}
         \vspace{-5mm}
         \caption{Real flight tracking of a circle trajectory in the $\mathrm{x}$, $\mathrm{y}$, and $\mathrm{z}$ axes.
         Left: position tracking performance; Right: linear velocity tracking performance.}
         \label{fig:cicrcle_trajectory-xyz}
      \end{figure*}

      We evaluate the performance of our policy on the real Crazyflie quadrotor to follow
      a circle trajectory start at $\mathrm{x}$ = $1$ and $\mathrm{y}$ = $0$ with a radius 
      of $1\mathrm{m}$ in $T$=$6$ seconds. 
      The formula of the circle trajectory is defined as:

      \begin{align}
         \label{eqn:circle_trajectory}
         p^{d}(t) 
         = \begin{bmatrix}
            x^{d}_{t} \\
            y^{d}_{t} \\
            z^{d}_{t} \\
         \end{bmatrix}  
         = \begin{bmatrix}
            1 \\
            0 \\
            1 \\
         \end{bmatrix}
         = \begin{bmatrix}
            cos(2\pi t/T) \\
            sin(2\pi t/T) \\
            0 \\
         \end{bmatrix}
      \end{align}
      where $T$ is the cycle time.

      The result of tracking performance is demonstrated in Fig. \ref{fig:circle_trajectory}.
      There is a minor tracking error but it is not a significant amount.
      The trained policy is able to track circle trajectories in $6$ seconds as in 
      Fig. \ref{fig:cicrcle_trajectory-xyz} where it reaches up to $1.8 \mathrm{m/s}$.
      The real world experiments video can be found at: \url{http://bit.ly/gnc\_drl\_quad\_controller}.

      We compared our proposed approach with different traditional controllers, including 
      Proportional-Integral-Derivative (PID) and Mellinger \cite{mellinger2011minimum}, 
      utilizing same trajectory tracking task, hardware configuration, and Crazyswarm default gains.
      Compared to the other controllers our trained controller directly generates RPMs as output 
      and hence does not take advantage battery voltage compensation, so we consider both the 
      Root-Mean-Square Error (RMSE) with and without the $\mathrm{z}$
      component as $\bar{e}$ and $\bar{e}_{xy}$ shown in Table \ref{tb3}, which are calculated as:
      \begin{align}
         \bar{e} &= \sqrt{\frac{1}{3} \left((x - x^{d})^{2} + (y - y^{d})^{2} + (z - z^{d})^{2}\right) } 
      \end{align}
      \vspace{-3mm}
      \begin{align}
         \bar{e}_{xy} &= \sqrt{\frac{1}{2} \left((x - x^{d})^{2} + (y - y^{d})^{2}\right)}
      \end{align}
      where $x^{d}$, $y^{d}$, $z^{d}$ are the desired position, 
      and $x$, $y$, $z$ are the actual position.

      Our controller policy has the capability of taking-off from the ground, which means overcoming 
      the impact of near-ground airflow effects.
      While tracking the given trajectory, the Mellinger controller outperforms the PID 
      controller due to the inclusion of an integral component, which aids in reducing 
      steady state error.

      Although our policy does not have integral part memory, but it is equivalent to 
      the Mellinger controller in terms of distance with $0.18\mathrm{m}$ and has a smaller error in 
      the $\mathrm{xy}$-plane with $0.16\mathrm{m}$. 
      \begin{table}[t!]
         \centering
         \caption{Comparison of tracking error when tracking the circle trajectory of different controllers.}
         \label{tb3}
         \resizebox{0.7\columnwidth}{!}{%
         \begin{tabular}{c c c}
            \toprule[1.5pt]
            \multirow{2}{*}{\textbf{Controller}} & \multicolumn{2}{c}{\textbf{Tracking error (m)}}   \\
                                             & $\bar{e}$             & $\bar{e}_{xy}$   \\
            \midrule[0.7pt]
            PID                                    & 0.68                  & 0.67                 \\
            Mellinger \cite{mellinger2011minimum}  & \textbf{0.18}         & 0.18                 \\
            \textbf{Ours}                          & \textbf{0.18}         & \textbf{0.16}                 \\
            \bottomrule[1.5pt]
         \end{tabular}%
      }
      \end{table}

\vspace{-5mm}
\section{Conclusion} 
\label{Conclusion}

   This research presents a reinforcement learning scheme for a quadrotor controller
   that generates RPMs directly based on the vehicle's states. Once trained with 
   the simulator on a computer, the policy is directly utilized on real-world 
   Crazyflie platforms for carrying out real flying tests in the VICON 
   positioning system, without necessity for expert parameter adjustments.
   The presented technique has been evaluated by experimental results, which 
   indicate its efficiency and practicality. The trained policy is able of 
   operating onboard and demonstrates precise trajectory tracking capabilities.

   There are still potential to further development in order to achieve even 
   more outstanding results. In future works, it is essential to enhance the 
   robustness by extending the policy to take into account the changes in system 
   parameters or environmental disturbances, such as variations in battery levels 
   or wind gusts.



%
\balance
\vspace{-5mm}
\bibliographystyle{ieeetr}
\bibliography{references}





%

\end{document}